\documentclass[9pt,conference]{IEEEtran}
\IEEEoverridecommandlockouts
\usepackage{amsmath,amssymb,amsfonts}
\usepackage{amssymb}
\usepackage[dvipsnames]{xcolor}
\usepackage{mathtools}
\usepackage{multirow}
\usepackage{lipsum}

\usepackage[normalem]{ulem}
\definecolor{ao(english)}{rgb}{0.0, 0.5, 0.0}

\usepackage{color,soul}
\usepackage{booktabs,makecell} % For formal tables
\usepackage{tabularx}

\usepackage{algorithm}
\usepackage{algpseudocode}

\usepackage{array, booktabs}
\usepackage{hyperref}
\newcolumntype{C}[1]{>{\centering\arraybackslash}p{#1}}

\def\BibTeX{{\rm B\kern-.05em{\sc i\kern-.025em b}\kern-.08em
    T\kern-.1667em\lower.7ex\hbox{E}\kern-.125emX}}

\usepackage[symbol]{footmisc}
\newcommand\blfootnote[1]{%
  \begin{NoHyper}
  \renewcommand\thefootnote{}\footnotetext{#1}%
  \addtocounter{footnote}{-1}%
  \endgroup
}

\usepackage{array}
\newcolumntype{P}[1]{>{\centering\arraybackslash}p{#1}}
\newcolumntype{L}[1]{>{\raggedright\arraybackslash}p{#1}}

\begin{document}

%%%%%%%%% TITLE
\title{Mamba Fusion: Learning Actions Through Questioning}
\author{\IEEEauthorblockN{Apoorva Beedu\IEEEauthorrefmark{1}\IEEEauthorrefmark{2}, Zhikang Dong\IEEEauthorrefmark{1}\IEEEauthorrefmark{2}\IEEEauthorrefmark{3}, Jason Sheinkopf \IEEEauthorrefmark{2}, Irfan Essa \IEEEauthorrefmark{2}\IEEEauthorrefmark{4}}
\IEEEauthorblockA{\IEEEauthorrefmark{2}Georgia Institute of Technology
\IEEEauthorrefmark{3}Stony Brook University
\IEEEauthorrefmark{4}Google Research}
\IEEEauthorblockA{\{abeedu3, zdong302, jsheinkopf, irfan\}@gatech.edu}
}

\maketitle

\begin{abstract}
\blfootnote{$ ^\star$ Denotes equal contribution.}
Video Language Models (VLMs) are crucial for generalizing across diverse tasks and using language cues to enhance learning. 
While transformer-based architectures have been the de facto in vision-language training, they face challenges like quadratic computational complexity, high GPU memory usage, and difficulty with long-term dependencies.
To address these limitations, we introduce \textbf{MambaVL}, a novel model that leverages recent advancements in selective state space modality fusion to efficiently capture long-range dependencies and learn joint representations for vision and language data.
{\textit{MambaVL} utilizes a shared state transition matrix across both modalities, allowing the model to capture a more comprehensive understanding of the actions in the scene.}
Furthermore, we propose a question-answering task that helps guide the model toward relevant cues. 
These questions provide critical information about actions, objects, and environmental context, leading to enhanced performance.
As a result, \textit{MambaVL} achieves state-of-the-art performance in action recognition on the Epic-Kitchens-100 dataset and outperforms baseline methods in action anticipation. 
The code is available at \url{https://github.com/Dongzhikang/MambaVL}.\\
\end{abstract}


%%%%%%%%% BODY TEXT

\section{Introduction}
\label{sec:intro}

Learning visual representations using paired vision-language descriptions have proven to be a powerful tool in computer vision. 
Natural language supervision \cite{pramanick2023egovlpv2,zhao2023learning} has advanced state-of-the-art performance not only in visual tasks such as classification~\cite{radford2021learning,pramanick2023egovlpv2} and object detection~\cite{minderer2022simple}, but also in other domains, including audio~\cite{wu2022wav2clip,dong2024musechat,liu2024tackling}, wearable sensor applications~\cite{haresamudram2024limitations}, and safety~\cite{jin2022towards,fu2024clipscope}.
Current video-language datasets are typically categorized as third-person or egocentric datasets. 
While third-person view data has been extensively studied, existing methods struggle to adapt to egocentric videos due to their unique characteristics, such as frequent and rapid camera motion, limited field of view, and occlusions caused by the user's body and objects.
However, the developments of large scale egocentric datasets such as Ego4D~\cite{grauman2022ego4d} and Epic-Kitchens-100~\cite{damen2022rescaling}, enable deeper exploration of this challenging setting.

Visual Language Models (VLMs) mainly rely on descriptions or captions, typically describing the contents of the videos \cite{ma2022x}.
This results in capabilities like cross-modal retrieval and zero shot recognition across diverse classification tasks \cite{ma2022x, zhao2023learning}. 
In contrast, we propose to \textit{generate two questions} -- one for the verb and one for the noun, each capturing distinct contextual information. 
This allows the model to better grasp the context of actions in the video leading to enhanced representation learning as questions encourage deeper reasoning and understanding relative to captions that describe the video content.
The advantage of such a setup are illustrated using the following example: Imagine a video showing the action ``drinking water'' and you're playing a guessing game where you must ask questions to help identify the action. 
To predict the verb and noun, two possible questions could be: \emph{(1)} ``What should you consume to quench your thirst?'' and \emph{(2)} ``What should you do to the water to enjoy its refreshing taste?''.
The answers to these questions reveal the action as a verb-noun pair within the scene. 
This approach mirrors how we often guide children toward answers by prompting them with questions, rather than simply providing the solution. 
Driven by this insight, we investigate if a question-driven framework can enhance the task of action recognition in the Epic-Kitchens-100 dataset.

{Most Transformer based video-language modeling can be broadly categorized into contrastive \cite{radford2021learning,beedu2024efficacy}, fusion-based learning~\cite{tang2023perceiver}, or a combination of both~\cite{pramanick2023egovlpv2}.}
While they are highly effective, the underlying attention mechanism requires quadratic computational complexity as the number of tokens grows. 
This limits their efficiency during both training and inference, and reduces their effectiveness in handling long-term sequence learning.
More recently however, Mamba, a state-space model based approach has emerged as an effective alternative, providing higher efficiency with linear scaling and reduced complexity \cite{gu2023mamba}.

In this work, we introduce \textbf{MambaVL}, a novel selective state space fusion model designed to process multi-modal input while efficiently handling long sequence modeling. 
The model employs a shared state transition matrix within the selective state space models (SSMs) for both the vision and language branches. 
This shared structure enables the SSMs to exchange information across modalities, extending the selection mechanism from single-modality models to multiple modalities. 
This approach allows the model to effectively \textit{select} relevant information across both vision and language domains, ensuring effective fusion -- which is highly flexible, and capable of integrating with any number or type of input modalities,  seamlessly incorporating existing pre-trained models.

Our contributions are: \emph{(1)} We propose a new question-answering task for recognizing actions in egocentric videos; \emph{(2)} We propose a novel vision-language fusion approach based on the selective state-space model, which can be straightforwardly adapted to handle inputs from any modality; and \emph{(3)} We conduct extensive experimental evaluation on the Epic-Kitchens-100 dataset \cite{damen2022rescaling}, achieving state-of-the-art performance on action recognition and action anticipation.

\begin{figure*}
    \centering
    \includegraphics[width=0.9\linewidth]{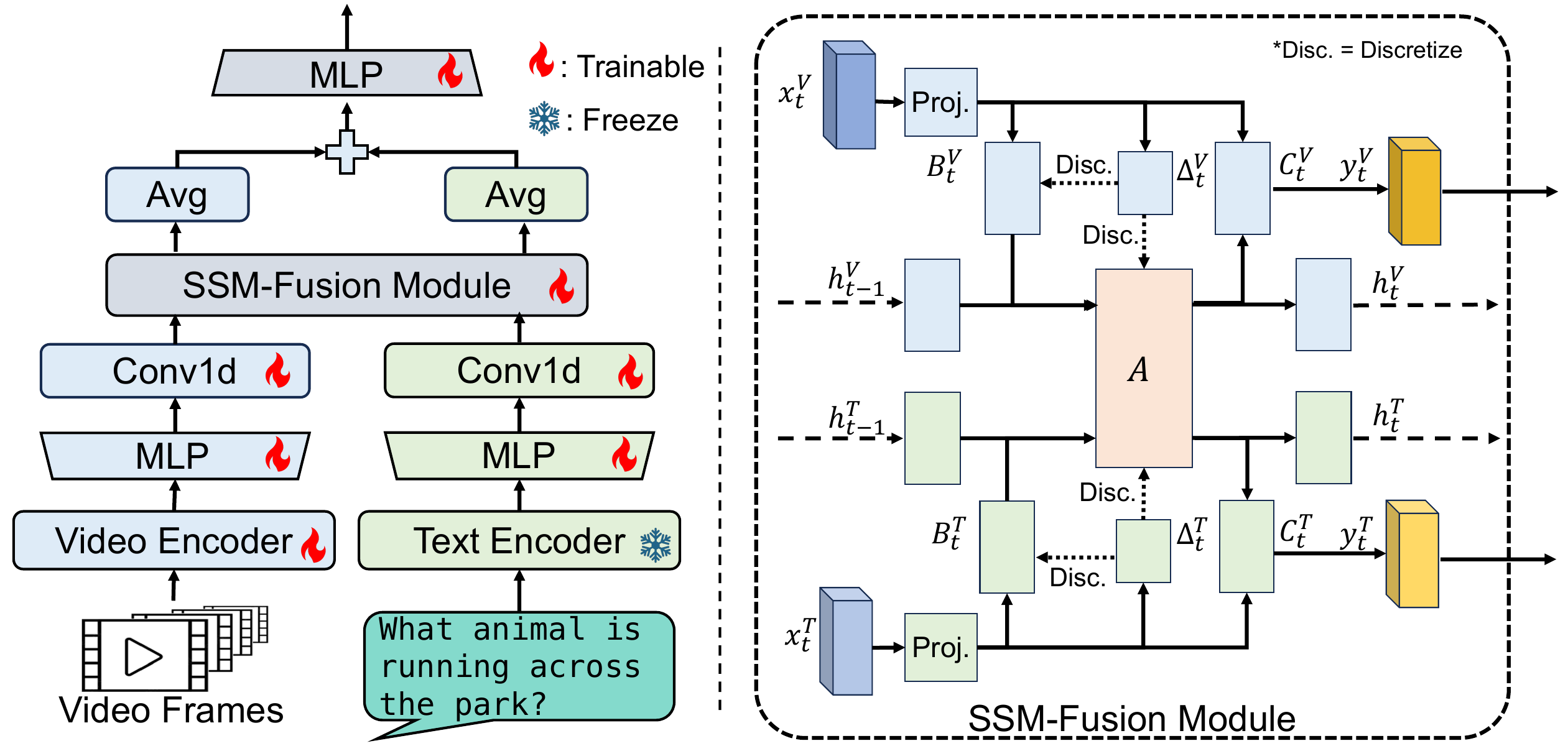}
    \caption{Overview of the SSM-Fusion module}
    \label{fig:overview}
\end{figure*}
\section{Related Work}
\label{sec:related}
\textbf{Vision-Language Models}. VLMs have seen rapid expansion, primarily driven by breakthroughs in image-language pre-training~\cite{radford2021learning,ma2022x,pramanick2023egovlpv2,girdhar2022omnivore}. 
Primarily, they are based on Transformers~\cite{vaswani2017attention}, either employing dual encoders that project visual and textual representations into a shared space to minimize distance~\cite{radford2021learning,ma2022x}, or shared encoders that concatenate features and feed them into a Transformer model~\cite{beedu2024efficacy,alamri2022end}. 
Recent works have also explored efficient approaches for learning representations for long-range video by utilizing cached memory~\cite{wu2022memvit}, through FlashAttention~\cite{zhao2023training}, or simplifying architectures by removing non-essential components~\cite{ryali2023hiera}.

\noindent\textbf{State Space Models in Computer Vision}. Structured State-Space Sequence model (S4) \cite{gu2021efficiently} is an efficient and effective method for modeling long-sequences, with linear scaling that has been applied in vision. 
\cite{islam2022long} utilizes S4 to learn temporal information for video classification. 
On the other hand, \cite{nguyen2022s4nd} introduces a multidimensional low-pass band-limiting S4 model to learn smooth convolutional kernels for images and videos. 
Going beyond, \cite{gu2023mamba} enhances S4 with a selectivity mechanism and proposes ``Mamba'', a promising alternative to Transformer models for capturing long-range dependencies. 
Originally not conceived for vision applications, recent works have begun to explore its potential in vision as well. 
For example, \cite{zhu2024visionmamba} learns visual representations through a bidirectional Mamba structure, while \cite{li2024videomamba} leverages Mamba to address challenges related to local redundancy and global dependencies in video understanding.

\noindent\textbf{Mamba for multi-modality fusion}. 
While Mamba \cite{gu2023mamba} can effectively process long-range sequences, it lacks a mechanism like cross-attention to learn representations from different modalities. 
Recent works have explored solutions to address this limitation. 
\cite{li2024mambadfuse} introduces a Mamba-based dual-phase fusion method to integrate complementary information across modalities. 
Similarly, \cite{dong2024fusion} develops a Fusion-Mamba block that maps various visual feature fusion blocks into a hidden state space. 
Different from the existing methods, we build a fusion model by sharing a common matrix -- the state transition matrix, while keeping the projection matrices modality specific. 
This helps the model to transfer information between different modalities, while also learning modality specific information.

\noindent\textbf{Video action recognition and anticipation}. An important requirement for these tasks is the ability to effectively capture long term temporal information and representation of objects in the video. 
While~\cite{patrick2021keeping,wu2022memvit} implicitly learn the motion path, ~\cite{herzig2022object} build on it to incorporate object representations. 
Other works, such as~\cite{girdhar2022omnivore,ma2022x,zhao2023training,zhao2023learning} learn generic long-range video representation by training on large datasets and fine-tune the model for downstream action related tasks. 
In contrast, we explore using natural language to learn object representations by formulating it as a question-answering task, and use selective state space models to learn joint representations.

\section{Methodology}
\label{sec:method}
Given the multi-modal inputs, the fusion model aims to combine and integrate the information from all modalities effectively. 
We present MambaVL, a vision-language fusion model that uses state-space models to learn joint representations from multiple modalities in~\autoref{fig:overview}.  

\begin{figure*}
    \centering
    \includegraphics[width=1\textwidth]{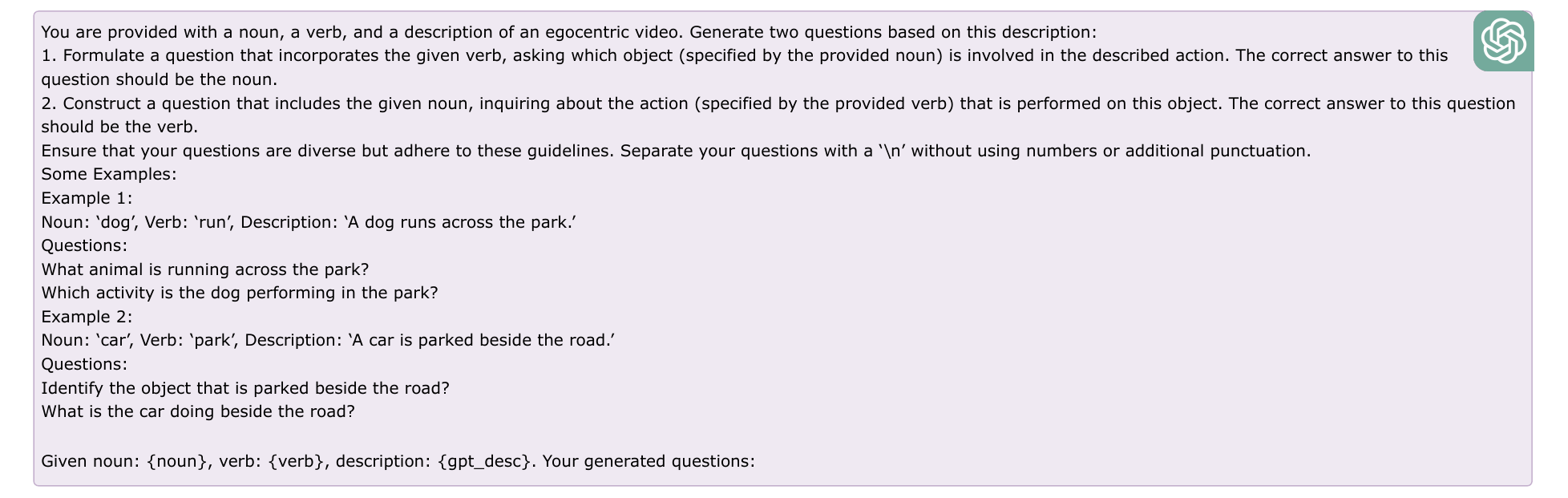}
    \caption{Prompt provided to ChatGPT to generate questions for the verb and the noun in action.
    }
    \label{fig:prompt}
\end{figure*}

\subsection{Preliminaries}
Consider a continuous system \eqref{eq:s4_definition}, which maps a 1-dimensional function or sequence $x(t) \in \mathbb{R} \mapsto y(t) \in \mathbb{R}$.
\begin{equation}
\begin{aligned}
h^{\prime}(t) & =\mathbf{A} h(t)+\mathbf{B} x(t), \\
y(t) & =\mathbf{C} h(t),
\end{aligned}
\label{eq:s4_definition}
\end{equation}
% \begin{equation}
% h^{\prime}(t) = \mathbf{A} h(t) + \mathbf{B} x(t), \quad y(t) = \mathbf{C} h(t),
% \label{eq:s4_definition}
% \end{equation}
where $\mathbf{A} \in \mathbb{R}^{\mathrm{N} \times \mathrm{N}}$, $\mathbf{B} \in \mathbb{R}^{\mathrm{N} \times \mathrm{1}}$, $\mathbf{C} \in \mathbb{R}^{\mathrm{1} \times \mathrm{N}}$ are the state transition, input projection and output projection matrices, respectively.

Building on this, \cite{gu2021efficiently} proposes the structured state space sequence (S4) model defined in the discrete space, where given a timescale parameter $\Delta$, the continuous matrices $\mathbf{A}$ and $\mathbf{B}$ are transformed into discrete matrices $\overline{\mathbf{A}}$ and $\overline{\mathbf{B}}$. The discretization process is zero-order hold (ZOH), which has the following form:
\begin{equation}
\begin{aligned}
& \overline{\mathbf{A}}=\exp (\boldsymbol{\Delta} \mathbf{A}), \\
& \overline{\mathbf{B}}=(\boldsymbol{\Delta} \mathbf{A})^{-1}(\exp (\boldsymbol{\Delta} \mathbf{A})-\mathbf{I}) \cdot \boldsymbol{\Delta} \mathbf{B}.
\end{aligned}
\label{eq:zoh}
\end{equation}

Using \eqref{eq:zoh}, \eqref{eq:s4_definition} can be rewritten as:
\begin{equation}
\begin{aligned}
h_t & =\overline{\mathbf{A}} h_{t-1}+\overline{\mathbf{B}} x_t, \; \;
y_t =\mathbf{C} h_t.
\end{aligned}
\label{eq:discrete_s4}
\end{equation}
By performing a global convolution in \eqref{eq:discrete_s4}, we have
\begin{equation}
\begin{aligned}
\overline{\mathbf{K}} & =\left(\mathbf{C} \overline{\mathbf{B}}, \mathbf{C} \overline{\mathbf{A B}}, \ldots, \mathbf{C A}^{k-1} \overline{\mathbf{B}}\right), \; 
\mathbf{y} =\mathbf{x} * \overline{\mathbf{K}},
\end{aligned}
\label{eq:global_conv}
\end{equation}
where $\overline{\mathbf{K}} \in \mathbb{R}^{\mathrm{k}}$ is a structured convolutional kernel.

\begin{algorithm}
\caption{SSM-Fusion Module}\label{algo:ssm-fusion}
\begin{algorithmic}[1]  % Add [1] to enable line numbering
\Require $\mathbf{x}^V$: \textcolor{red}{$(\mathrm{B}, \mathrm{F}, \mathrm{D})$}, $\mathbf{x}^T$: \textcolor{red}{$(\mathrm{B}, \mathrm{L}, \mathrm{D})$}
\Ensure $\mathbf{y}^V$: \textcolor{red}{$(\mathrm{B}, \mathrm{F}, \mathrm{D})$}, $\mathbf{y}^T$: \textcolor{red}{$(\mathrm{B}, \mathrm{L}, \mathrm{D})$}
\State $\boldsymbol{A}$: \textcolor{red}{$(\mathrm{D}, \mathrm{N})$} $\leftarrow$ Parameter
\State $\boldsymbol{B}^V$: \textcolor{red}{$(\mathrm{B}, \mathrm{F}, \mathrm{N})$} $\leftarrow$ Linear$_\mathrm{B}^{V}(\mathbf{x}^V)$
\State $\boldsymbol{B}^T$: \textcolor{red}{$(\mathrm{B}, \mathrm{L}, \mathrm{N})$} $\leftarrow$ Linear$_\mathrm{B}^{T}(\mathbf{x}^T)$
\State $\boldsymbol{C}^V$: \textcolor{red}{$(\mathrm{B}, \mathrm{F}, \mathrm{N})$} $\leftarrow$ Linear$_\mathrm{C}^{V}(\mathbf{x}^V)$
\State $\boldsymbol{C}^T$: \textcolor{red}{$(\mathrm{B}, \mathrm{L}, \mathrm{N})$} $\leftarrow$ Linear$_\mathrm{C}^{T}(\mathbf{x}^T)$

\State $\Delta^V$: \textcolor{red}{$(\mathrm{B}, \mathrm{F}, \mathrm{D})$} $\leftarrow$ Softplus$($Parameter + Linear$_\mathrm{\Delta}^{V}(\mathbf{x}^V)$$)$
\State $\Delta^T$: \textcolor{red}{$(\mathrm{B}, \mathrm{L}, \mathrm{D})$} $\leftarrow$ Softplus$($Parameter + Linear$_\mathrm{\Delta}^{T}(\mathbf{x}^T)$$)$

\State $\overline{\boldsymbol{A}^V}, \overline{\boldsymbol{B}^V}$: \textcolor{red}{$(\mathrm{B}, \mathrm{F}, \mathrm{D}, \mathrm{N})$} $\leftarrow \text{Discretize}(\Delta^V, \boldsymbol{A}, \boldsymbol{B}^V)$
\State $\overline{\boldsymbol{A}^T}, \overline{\boldsymbol{B}^T}$: \textcolor{red}{$(\mathrm{B}, \mathrm{L}, \mathrm{D}, \mathrm{N})$} $\leftarrow \text{Discretize}(\Delta^T, \boldsymbol{A}, \boldsymbol{B}^T)$

\State $\mathbf{y^V}$: \textcolor{red}{$(\mathrm{B}, \mathrm{F}, \mathrm{D})$} $\leftarrow$ SSM$(\overline{\boldsymbol{A}^V}, \overline{\boldsymbol{B}^V},C^V)$
\State $\mathbf{y^T}$: \textcolor{red}{$(\mathrm{B}, \mathrm{L}, \mathrm{D})$} $\leftarrow$ SSM$(\overline{\boldsymbol{A}^T}, \overline{\boldsymbol{B}^T},C^T)$ \\

\Return $\mathbf{y}^V$ and $\mathbf{y}^T$
\end{algorithmic}
\end{algorithm}

\subsection{MambaVL}
\autoref{fig:overview} shows the overall structure of MambaVL. 
\cite{gu2023mamba} introduced ``Mamba" block to process the single modality sequence data, where, the timescale parameter $\Delta_t$, ${\mathbf{B}}$ and $\mathbf{C}$ are directly derived from the input. 
As $\mathbf{A}$ is not directly dependent on the input, we hypothesize that sharing the state transition matrix $\mathbf{A}$ between different modalities enables the model to learn joint representations. 

Given a video clip $\mathbf{V} \in \mathbb{R}^{F\times C\times H\times W}$, with $F$ frames, $C$ channels, and $(H,W)$ frame size, we use a video encoder to extract video features, denoted as $\mathbf{x}^V$, and the RoBERTa model~\cite{liu2019roberta} to extract embeddings from raw text input (denoted as $\mathbf{x}^T$).
We use MLP layers to project video and text embeddings into a shared space, followed by 1-D convolution layers on each embedding type.
\begin{equation}
\begin{aligned}
\mathbf{x}^V = \text{Conv1d(MLP(}\mathbf{x}^V\text{))}, \\
\quad \mathbf{x}^T = \text{Conv1d(MLP(}\mathbf{x}^T\text{))}.
\end{aligned}
\label{eq:mlp_conv1d}
\end{equation}
where $\mathbf{x}^V = \left\{x_1^V, \ldots, x_F^V \right\}$, $\mathbf{x}^T = \left\{x_1^T, \ldots, x_L^T\right\}$, and $L$ is the sequence length of the text tokens.

We detail our method in Algorithm \ref{algo:ssm-fusion}. 
We use $\mathbf{x}^V$ and $\mathbf{x}^T$ as input to our SSM-Fusion Module, and initialize a shared state transition matrix $\boldsymbol{A}$ for two modalities. 
Then, we utilize linear projection layers to transform $\mathbf{x}^V$ and $\mathbf{x}^T$ into $\boldsymbol{B}^V, \boldsymbol{B}^T, \boldsymbol{C}^V, \boldsymbol{C}^T, \Delta^V \text{ and } \Delta^T$ respectively. 
By applying different selective time scale parameters in Eqn.\ \eqref{eq:zoh}, we obtain parameter triplets $(\overline{\boldsymbol{A}^V}, \overline{\boldsymbol{B}^V}, \boldsymbol{C}^V)$ and $(\overline{\boldsymbol{A}^T}, \overline{\boldsymbol{B}^T}, \boldsymbol{C}^T)$. 
We compute the SSM in Eqn.\ \eqref{eq:discrete_s4} to get the output $\mathbf{y}^V$ and $\mathbf{y}^T$. 
Finally, we perform average pooling at the sequence level and sum the two outputs to get the overall fused representation.
These fused representations are then passed through a linear layer for classifications tailored to specific tasks.

\noindent\textbf{Interpretation of the shared matrix} $\boldsymbol{A}$. The shared state transition matrix $\boldsymbol{A}$ serves as a ``cross modality selective'' mechanism for information exchange between modalities. 
It aids in discretizing matrix $\boldsymbol{B}$ and in computing the output. 
This enables each modality branch to adjust its recurrent dynamics by integrating the current input, hidden states, and cross-modal information flow. 
During backpropagation, matrix $\boldsymbol{A}$ is updated using gradient from all modalities, allowing it to comprehensively capture information across the modalities. 
% Generally, this mechanism can be extended to any combination of modalities.
This shared mechanism enables our model to unify temporal dynamics across modalities, as the state transition matrix  $A$  governs the temporal evolution of states, as shown in Equation \eqref{eq:s4_definition}.
Additionally, using a shared matrix facilitates the learning of interdependency between different modalities by enabling gradient flow across them.
\section{Experiments}
\label{sec:exp}
\subsection{Experimental Setup}
\textbf{Datasets and Metrics: } 
We evaluate our approach on the Epic-Kitchens-100 (EK100) dataset on two tasks: \textit{Action recognition and Action Anticipation.}
\textbf{EK100}~\cite{damen2022rescaling} is an egocentric video dataset capturing daily activities around the kitchen.
Action anticipation and classification task requires each video classifying/predicting one of the 97 verbs and 300 nouns.
The network’s highest-scoring verb-noun pair defines the action.
We report the top-1 accuracy for action recognition and Recall@5 for action anticipation.

\textbf{Implementation Details:} We set the input frame length to 16 frames for all our experiments, and follow the same data processing pipelines as the specific backbones used. 
Training is performed for 100 epochs, where we employ cosine annealing with a warmup for $2$ epochs, with a base learning rate of $1e^{-6}$, which linearly increases to a peak of $1e^{-3}$, and then gradually decreases to $1e^{-5}$ following a half-wave cosine schedule..
The Mamba block has a higher peak learning rate, set to $3e^{-3}$. 
We use a batch size of 128, distributed over 8 Nvidia A40 GPUs.
For action anticipation, we set the anticipation time $\tau_{a}$ as 1 second, and use 16-frame long clips at a resolution of 224 × 224, uniformly sampled from an observed video of 64 frames ($\sim$2s in total). 
For action anticipation, we use the same configuration as ORViT\footnote[1]{For dataloader and metrics, we follow \href{https://github.com/LAHAproject/InAViT/tree/main}{InAViT}} to train the baseline model. 
We use AVION~\cite{zhao2023learning} to recognize actions in the observed frames and use these \textit{recognized actions} to generate questions for the action anticipation task. 

\textbf{Question Generation:} We introduce a new annotation framework consisting of question-answer pairs for actions in the EK100 dataset. 
This process is divided into two stages: 
first, the relatively simple narrations in the EK100 dataset is rephrased and enriched with detailed action descriptions using the GPT-4o~\cite{openai2024_gpt4o} to cope with the lack of diversity in the original narrations.
In the second stage, using the verb, noun, and action description for each sample, we prompt GPT-4o to generate two distinct questions for each action: one targeting the verb and the other the noun, as illustrated in~\autoref{fig:prompt}.
To prevent data leak during training, where the task is to predict verbs and nouns simultaneously, we mask the verb in the noun-related question with a \texttt{<MASK>} token if it appears, and vice versa for the noun in verb-related questions.
In our final dataset, each sample comprises two questions and their corresponding answers.

\subsection{Results and Analysis}
 \textbf{Action Recognition:}
In~\autoref{tab:ek100_cls}, we compare MambaVL against state-of-the-art methods for action recognition.
We further evaluate the effects of model size by employing three different visual encoder backbones: ORViT, ViT-B and ViT-L.
Consistently, MambaVL outperforms \textit{all baseline methods} for actions, verb and noun classification. 
MambaVL also outperforms LaViLa~\cite{zhao2023learning}, another vision-language method, indicating the effectiveness of mamba-based fusion of modalities.
MambaVL outperforms LaViLa by $>3\%$ for both the base and the large version, and 3\% improvement over ORViT. 
    As reported in~\autoref{tab:fusion_flops}, our mamba based fusion block adds negligible number of FLOPs and trainable paramters, while providing a significant improvement in accuracy.

 \textbf{Action Anticipation:}
We also evaluate MambaVL for the task of action anticipation against baselines like AVT~\cite{girdhar2021anticipative}, AFFT~\cite{zhong2023anticipative}, MeMViT~\cite{wu2022memvit} and ORViT~\cite{herzig2022object}. 
We trained ORViT for action anticipation using their publicly available code, while using the same training configurations as MambaVL.
We do not compare against other LLM based methods as our model does not train/finetune an LLM, instead uses GPT-4 to generate the questions.
{While we outperform ORViT's baseline results, our performance is somewhat constrained by AVION, as we rely on the detected actions to generate relevant questions. 
This introduces potential errors into our prediction pipeline, which can be mitigated when a more accurate action recognition model is used.}

\begin{table}[!t]
    \centering
        \caption{Comparison of the state-of-the-art for action recognition on EK100. We report Top-1 \% for verb, noun and action classification. }
    \setlength{\tabcolsep}{2pt}
    \resizebox{1\columnwidth}{!}{
    \begin{tabular}{C{.33\columnwidth} C{.4\columnwidth} C{0.1\columnwidth} C{.1\columnwidth} C{.1\columnwidth}}
    \toprule
    Model(Backbone) & Pretrain data & Verb & Noun & Action \\ \midrule
    MeMViT (24x3) & K600 & 71.4 & 60.3 & 48.4 \\
    Omnivore (swin-B) & \small{IN-(21k+1k)+K400+SUN} & 69.5 & 61.7 & 49.9 \\ \midrule
    MeMViT (16x4) & K400 & 70.6 & 58.5 & 46.2 \\
    ORViT (MF-HR) & IN-21k+K400 & 68.4 & 58.7 & 45.7 \\
    \textbf{MambaVL (ORViT)} & IN-21k+K400 & \textbf{69.1} & \textbf{63.9} & \textbf{48.6} \\ \midrule
    AVION (ViT-B) & WIT + Ego4D & 70.0 & 59.8 & \textbf{49.1} \\
    LaViLa (TSF-B) & WIT + Ego4D & 69.0 & 58.4 & 46.9 \\
    \textbf{MambaVL (ViT-B)} & WIT + Ego4D & \textbf{70.9} & \textbf{61.1} & \textbf{49.1} \\ \midrule
    AVION (ViT-L) & WIT + Ego4D & 73.0 & 65.4 & 54.4 \\
    LaViLa (TSF-L) & WIT + Ego4D & 72.0 & 62.9 & 51.0 \\ 
    \textbf{MambaVL (ViT-L)} & WIT + Ego4D & \textbf{74.3} & \textbf{67.1} & \textbf{55.0} \\ 
    \bottomrule
    \end{tabular}
    }
    \label{tab:ek100_cls}
\end{table}

\begin{table}[t]
\centering
\caption{Model comparison by GFLOPS and parameter count.}
\setlength{\tabcolsep}{4pt}
\resizebox{0.75\columnwidth}{!}{
\begin{tabular}{ccc}
\hline
{Model}& {GFLOPS} & {Params} \\ \hline
ORViT& 405& 148M\\
ORViT + Transformer Fusion&413.5&242M\\
MambaVL& 413& 157M\\ \hline
\end{tabular}
}
\label{tab:fusion_flops}
\end{table}

\textbf{Impact of the Fusion Module}
In~\autoref{tab:fusion_ablation}, we compare the performance of Mamba-based fusion (i.e., MambaVL) against other fusion methods: \emph{(i)} 2-layer MLP, where the text embeddings and visual features from ORViT are concatenated and passed through the MLP; \emph{(ii)} a Transformer containing six layers and four heads; and \emph{(ii)} a Transformer with 12 layers and 12 heads. 
We see that our Mamba-based fusion outperforms other methods by a significant margins, by approx.\ 7-10\%, indicating that MambaVL is capable of encoding long range information while also learning effective joint representations.

\begin{table}[!t]
    \centering
        \caption{Comparison of the state-of-the-art for action anticipation on EK100. We report Top-1 \% for verb, noun and action prediction. }
    \setlength{\tabcolsep}{2pt}
    \resizebox{1\columnwidth}{!}{
    \begin{tabular}{C{.3\columnwidth} C{.3\columnwidth} C{0.1\columnwidth} C{.1\columnwidth} C{.1\columnwidth}}
    \toprule
    \multicolumn{1}{c}{\multirow{2}{*}{Method}} & \multicolumn{1}{c}{\multirow{2}{*}{Pretrain data}} & \multicolumn{3}{c}{Overall} \\ \cline{3-5} 
     & & Verb & Noun & Action \\ \midrule
    AVT+~\cite{girdhar2021anticipative} & IN21K + EPIC boxes & 28.2 & 32.0 & 15.9 \\
    MeMViT (32x3)~\cite{wu2022memvit} & K700 & \underline{32.2} & \textbf{37.0} & 17.7 \\
    MeMViT (16x4)~\cite{wu2022memvit} & K400 & \textbf{32.8} & 33.2 & 15.1 \\
    AFFT~\cite{zhong2023anticipative} & IN-21K& 22.8 & 34.6 & 18.5 \\
    ORViT-MF~\cite{herzig2022object} & IN-21k+K400 & 26.9 & 34.2 & \underline{23.3} \\ \midrule
    \textbf{MambaVL (ORViT) }& IN-21k+K400 & 29.1 & \underline{35.1} & \textbf{23.9} \\
    \bottomrule
    \end{tabular}
    }
    \label{tab:ek100_ant}
\end{table}

\begin{table}[!t]
    \centering
    \caption{Comparison between different fusion methods. }
    \setlength{\tabcolsep}{2pt}
    \resizebox{1\columnwidth}{!}{
    \begin{tabular}{C{.3\columnwidth} C{0.2\columnwidth} C{.2\columnwidth} C{.2\columnwidth}}
    \toprule
    \multicolumn{1}{c}{\multirow{2}{*}{Fusion Method}} & \multicolumn{3}{c}{Overall} \\ \cline{2-4} 
     & Verb & Noun & Action \\ \midrule
    MLP & 62.8 & 51.6 & 39.6 \\
    Transformer (6x4)  & 62.9 & 51.9 & 40.0 \\
    Transformer (12x12) & 62.5 & 51.8 & 39.5 \\ \midrule
    \textbf{MambaVL} & \textbf{69.1} & \textbf{63.9} &\textbf{ 48.6} \\
    \bottomrule
    \end{tabular}
    }
    \label{tab:fusion_ablation}
\end{table}

\section{Conclusion}
\label{sec:conclusion}
In this paper, we introduced MambaVL, a novel approach for fusing visual and language features. 
Our key innovation is the use of a learnable shared state transition matrix within the SSM block for each modality, which along with the selection mechanism, enables each modality to learn from its own input while also considering information from other modalities during training. 
This lightweight fusion method is not only flexible enough to accommodate any number of input modalities, but it is also compatible with a wide range of pretrained feature extraction models. 
We also introduced Question-Answering as a viable alternative for the video based classification tasks such as action recognition and anticpation for VLMs.
Our findings demonstrate the effectiveness of MambaVL in cross-modal fusion, and we believe it opens new opportunities for further research into Mamba's application in cross-modal tasks.

%%%%%%%%% REFERENCES
\bibliographystyle{IEEEtran}
\bibliography{sample-base}

\end{document}